\documentclass[10pt,twocolumn,letterpaper]{article}

\usepackage{cvpr}
\usepackage{times}
\usepackage{epsfig}
\usepackage{graphicx}
\usepackage{amsmath}
\usepackage{amssymb}
\usepackage{hyperref}
\usepackage{graphicx}
\usepackage{booktabs}
\usepackage{multirow}
\usepackage{caption}
\usepackage{enumerate}
\usepackage{subcaption}
\usepackage[shortlabels]{enumitem}
\usepackage{comment}
\usepackage{multicol}
\usepackage{lipsum}

\begin{document}

%%%%%%%%% TITLE
%\title{Evaluating PAD Algorithm Performance on Synthetic Live and Textured Contact lenses Irises: Mitigating Identity Leakage and Ethical Concerns}
\title{Privacy-Safe Iris Presentation Attack Detection}

\author{Mahsa Mitcheff, Patrick Tinsley, and Adam Czajka\\
384 Fitzpatrick Hall of Engineering, University of Notre Dame, IN 46556, USA\\
{\tt\small \{mmitchef,ptinsley,aczajka\}@nd.edu}
}

\newcommand{\teaser}{
{
    \vskip3mm
    \begin{center}
    \vskip-5mm\includegraphics[width=0.9\linewidth]{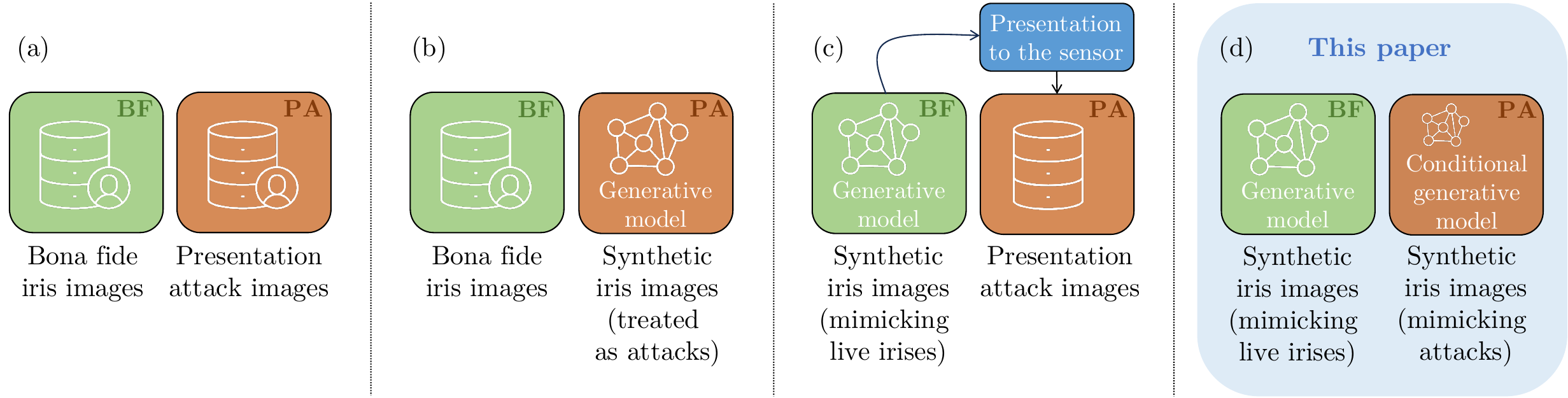}
    
    % Adam's version
    \captionof{figure}{Approaches to source training samples to develop iris presentation attack detection methods: (a) a classical approach, in which authentic images of living individuals and physical attacks are used, (b) similar to (a) but targeting specifically detection of synthetically-generated iris images, or ``deep fakes'', (c) a newer approach, in which {\it bona fide} samples are synthesized, and then used to carry out physical presentation attacks captured by iris sensors, and (d) the approach proposed in this paper, in which both {\it bona fide} and attack samples are synthesized and used to train iris presentation attack detection models. {\bf BF} and {\bf PA} stand for ``bona fide'' and ``presentation attack,'' respectively. A person icon next to the dataset icon denotes the presence of identity information in the data.}
    \label{fig:teaser}
    \end{center}
}
}

\maketitle
\thispagestyle{empty}

%%%%%%%%% ABSTRACT
\begin{abstract}
\vspace*{-20pt}
This paper proposes a framework for a privacy-safe iris presentation attack detection (PAD) method, designed solely with synthetically-generated, identity-leakage-free iris images. Once trained, the method is evaluated in a classical way using state-of-the-art iris PAD benchmarks. We designed two generative models for the synthesis of ISO/IEC 19794-6-compliant iris images. The first model synthesizes bona fide-looking samples. To avoid ``identity leakage,'' the generated samples that accidentally matched those used in the model's training were excluded. The second model synthesizes images of irises with textured contact lenses and is conditioned by a given contact lens brand to have better control over textured contact lens appearance when forming the training set. Our experiments demonstrate that models trained solely on synthetic data achieve a lower but still reasonable performance when compared to solutions trained with iris images collected from human subjects. This is the first-of-its-kind attempt to use solely synthetic data to train a fully-functional iris PAD solution, and despite the performance gap between regular and the proposed methods, this study demonstrates that with the increasing fidelity of generative models, creating such privacy-safe iris PAD methods may be possible. The source codes and generative models trained for this work are offered along with the paper. 

\end{abstract}

%%%%%%%%% BODY TEXT
\section{Introduction}
A textured contact lens (TCL), if worn, partially occludes the natural pattern of an iris, and thus almost certainly leads to a false non-match result if the iris covered by such a lens is being matched to the same, clear iris \cite{Doyle_ICB_2013,Yadav_TIFS_2014,Czajka_CSUR_2018}. 
This, combined with the fact that the true intention of wearing contact lenses cannot be easily assessed (\eg, cosmetic/medical reasons), has led to TCLs being one of the most popular concealer attacks on iris recognition systems in real operational scenarios. 
The quest for effective iris presentation attack detection (PAD) methods, specifically crafted for detecting TCLs, is thus unfading.

Effective iris PAD methods, especially those based on modern deep learning-based models, are data-hungry. Acquiring large sets of authentic iris images without and with TCLs (and of multiple and diverse brands) may become more challenging in the near future due to rising debate around Artificial Intelligence (biometrics included) and the ways of how our privacy is protected when our data is used to train AI methods \cite{Tinsley_WACV_2021, tinsley2022haven, jain2021biometrics}.

However, these fears can be transformed into opportunities.
With constantly increasing fidelity of image synthesis (including biometric samples) offered by modern generative models, we may eventually reach a point when authentic and synthetically-generated iris images will be indistinguishable. This may include synthetic iris images showing rare and difficult-to-collect anomalies, such as deformations caused by post-mortem decomposition \cite{bhuiyan2024forensic}, but also irises with TCLs of various and unknown brands. This paper investigates this latter opportunity and makes an attempt to train a regular iris PAD model using {\it solely} synthetically-generated data. Fig. \ref{fig:teaser} illustrates the difference between the proposed approach and other works related to iris PAD and utilizing synthetically-generated samples. Additionally, we also make sure that none of the synthetic iris images ``leak'' identity information from the training samples (used to train the generative models) to the synthetic data employed in training our PAD models. This is done by excluding samples synthesized by the generative model that are ``too close'' (in a biometric matching sense) to samples used in training this generative model. Such ``privacy-safe'' PAD models are then tested as usual, with regular iris PAD benchmarks containing authentic iris images with and without TCLs, and, interestingly, achieve performance which is not significantly worse than the one obtained for models trained with authentic data.

Our experiments show that the PAD accuracy of models trained solely with synthetic data is statistically significantly worse than models trained with the same number of samples but originating from human subjects: the average Area Under the ROC Curve (AUC) across five independent training runs and across all models equal to 0.90 (lowest) and 0.93 (highest) in the former case, compared to 0.97 obtained in the latter case. However, this discrepancy in accuracy is relatively small and demonstrates the feasibility of training iris PAD methods using only synthetic data. This accuracy gap may disappear quickly with increasing fidelity of biometric data synthesizers.

As a summary, the {\bf novel contributions} of this paper are:

\begin{enumerate}[start=1,label={\bfseries C\arabic*: }]
    \item a framework to train a privacy-safe iris PAD method with solely synthetically-generated data, 
    \item a proof of concept demonstrating current capabilities of the approach when tested on state-of-the-art iris PAD benchmarks,
    \item all elements needed to replicate our results and facilitate further research in this area: (a) source codes and weights of unconditional and conditional StyleGAN2-ADA models synthesizing ISO/IEC 19794-6-compliant iris images without and with textured contact lenses (mimicking seven distinct contact lens brands)\footnote{\url{https://github.com/CVRL/PrivacySafeIrisPAD}}, and (b) dataset of synthetic iris images used to train the privacy-safe PAD models\footnote{instructions of how to request a copy of the dataset can be found at \url{https://cvrl.nd.edu/projects/data/}; look for dataset named {\it Privacy-Safe Iris PAD (UND-2024-PSIPAD)}}.
\end{enumerate}

\section{Related Work}

\subsection{Iris Image Synthesis}

Generative Adversarial Networks (GANs)\cite{goodfellow2014generative} have revolutionized the field of image generation, enabling the creation of highly realistic synthetic images. 
Researchers have extensively used GAN-based models to also generate iris images. Such first attempts were made by Kohli \etal \cite{kohli2017synthetic} and Minaee and Abdolrashidi \cite{minaee2018iris}, who introduced the iris synthesis models based on deep convolutional generative adversarial network, called iDCGAN and DC-GAN, respectively. Yadav \etal \cite{yadav2019synthesizing} employed a relativistic average standard GAN (RaSGAN), which -- unlike iDCGAN -- trains its generator to maximize the probability that randomly sampled synthetic iris images are more realistic than a set of real irises. The problem with these early, yet successful, approaches was that generated images had the iris filling out the image frame almost entirely (we could classify this type as a {\it cropped} iris image, according to ISO/IEC 19794-6), and were consequently lacking some of the fine iris details due to low image resolution.

Image-to-image translation approaches were also found to be useful in iris image synthesis. Zou \etal \cite{zou2018generation} proposed 4DCycle-GAN that adds two more discriminators to Cycle-GAN to increase diversity of features in synthesized irises. Their goal was to convert irises with textured contacts lenses into images of irises without texture contact lenses. Yadav and Ross \cite{yadav2021cit} proposed a cyclic image translation generative adversarial network (CIT-GAN) for iris image manipulation. The model transfers stylistic elements from one domain to another domain, \eg from cosmetic contact lenses to plain iris. More recently, Yadav and Ross \cite{yadav2023iwarpgan} proposed the iWarpGAN model that can generate irises of new identities not seen during training, with the increased variety of image styles present in the generated irises.

The above works lack one important, from the biometrics point of view, property of the synthesized images: identity preservation. This means that generated samples, even exhibiting high visual realism, could not be considered as same-eye samples. This has been partially addressed by Khan \etal \cite{Khan_WACV_2023,khan2023eyepreserve}, who proposed deep learning-based iris pattern deformation models to synthesize iris images with different pupil sizes, correctly modeling the anatomical nonlinear deformations of the iris tissue, and thus preserving identity.
Bhuiyan and Czajka \cite{bhuiyan2024forensic} also propose a conditional StyleGAN2-based model to generate both same-eye and different-eye post-mortem iris images, given the time elapsed since death (post-mortem interval). 

\subsection{Iris Presentation Attack Detection}

Iris PAD has seen significant attention to date, with more than two hundred papers proposing iris PAD algorithms published to date, if we base our estimates solely on the methods surveyed by Czajka and Bowyer \cite{Czajka_CSUR_2018} and Boyd \etal \cite{Boyd_PRL_2020}, as well as algorithms submitted to the LivDet-Iris competitions \cite{yambay2023review,Tinsley_IJCB_2023}. Despite the increasing performance of iris PAD in closed set scenarios (when attack types used in testing are known), especially demonstrated by deep learning-based algorithms, iris PAD still lacks the ability to generalize well to unknown attack types and unknown properties of known attack types \cite{Boyd_TIFS_2023}. {\bf This paper addresses the latter challenge, in which properties} (brand, texture, opaqueness) {\bf of known attack type} (textured contact lenses) {\bf are unknown}.

\subsection{Synthetic Data in the Context of PAD}

The fidelity of modern generative models has increased so far that instead of treating generated images as spoof examples, synthesized images started to replace authentic samples. Yadav \etal \cite{yadav2019synthesizing} proposed an approach in which synthetically-generated irises were either treated as authentic samples, or were added to enlarge the set of presentation attack types. In both scenarios, the authors found that adding synthetic samples increases the generalization capabilities of iris PAD. Yadav \etal \cite{yadav2021cit} proposed a generative model to transfer the style of known presentation attacks (\eg, paper printouts or artificial eyes) onto a bona fide samples. 

Both solutions above may considerably increase the size of iris PAD-related datasets. One potential aspect not yet discussed is the increase of privacy: whether generative models are appropriately used in the PAD context. Fang \etal explicitly included the discussion about privacy into synthesizing PAD datasets \cite{fang2023synfacepad} and introduced {\it SynthASpoof}, the first privacy-preserving dataset for training face PAD algorithms \cite{fang2023synthaspoof}. This dataset consists of synthetically-generated faces (via StyleGAN2-ADA) that serve as ``bona fide'' samples. The researchers then captured real physical presentation attacks (printed and displayed images of synthetic faces), % SynthASpoof 
and demonstrated the feasibility of training face PAD models solely with synthetic data, eliminating the ethical and legal issues associated with collecting images of real faces. 
Later, {\it SynthASpoof} was used in the ``Face Presentation Attack Detection Based on Privacy-Aware Synthetic Training Data'' competition (SynFacePAD 2023) \cite{fang2023synfacepad}, demonstrating promising results achieved by the competition participants in training PAD models solely on synthetic face images from SynthASpoof. 

{\bf Our paper differs from the above works in at least two important aspects}. Firstly, both ``bona fide'' and ``spoof'' samples are synthesized by the proposed generative models. This eliminates the need to carry out physical presentation attacks, and thus naturally makes the creation of the PAD dataset much easier. Secondly, and more importantly, due to a potential data leakage observed in generative models, we have added a component to verify that synthesized samples do not biometrically match samples of authentic subjects, whose data was used in training the generators.

\section{Methodology}

\begin{figure*}[!th]
    \begin{center}
    \includegraphics[width=\linewidth]{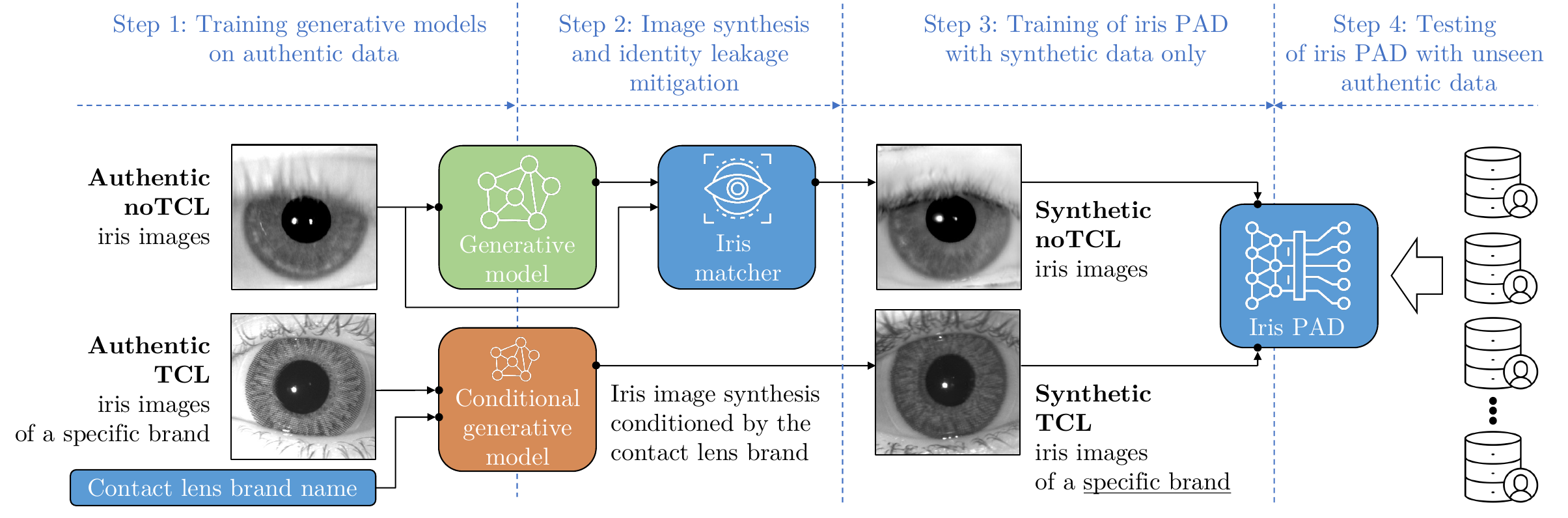}
    
    \captionof{figure}{The pipeline of privacy-safe, synthetic data-only iris presentation attack detection (PAD) training and validation. {\bf TCL} and {\bf noTCL} denote images of irises with and without contact lenses, respectively. After training generative models ({\bf Step 1}), we exclusively use synthetically-generated data (mimicking irises both with and without textured contact lenses) to train iris PAD as usual ({\bf Step 3}). The iris matcher is used (in {\bf Step 2}) to exclude synthetic samples that are ``too close'' to non-synthetic samples used for generative models training, which prevents the ``leakage'' of identity information from the training set into the generated samples. Resulting iris PAD methods are tested on regular (non-synthetic) data composed of {\it bona fide} and fake samples ({\bf Step 4}).}
    \label{fig:pipeline}
\end{center}
\end{figure*}

\subsection{Generative Model Selection}

Limited data per contact lens brand presents a challenge for choosing a suitable generative model. Conventional Generative Adversarial Networks (GAN)-based models typically require considerably large datasets. However, in the case of TCL iris images, we usually only have a few hundred to a few thousand images per lens brand, which may cause the discriminator to overfit. As such, we opted for the StyleGAN2-ADA architecture \cite{Karras2020ada}, which is designed for synthesizing images after training with limited number of samples. The StyleGAN2-ADA framework is designed to increase the number and variety of samples through differing augmentation techniques. 
The applied training data augmentations followed those recommended by the StyleGAN2-ADA authors.

It is important to note that the presented framework is not generative model-specific (for instance, StyleGAN-specific). In particular, the identity leakage mitigation component is a post-hoc operation once the synthesis is completed, and thus can be applied to any generator. In general, any generative model able to synthesize ISO-compliant iris images, such as diffusion models \cite{Rombach_CVPR_2022,SohlDickstein_ICML_2015}, can be deployed. 

\subsection{Overview of the Proposed Framework}

Fig. \ref{fig:pipeline} outlines a framework consisting of four distinct steps to train an iris PAD system while preserving user privacy.
% Adam's version:
To synthesize iris images, we trained two separate StyleGAN2-ADA-based models ({\bf Step 1} in Fig. \ref{fig:pipeline}). The first model synthesizes irises with TCLs, and is conditioned by the contact lens brand, enabling image generation with contact lenses matching designs and patterns of seven distinct brands: Bausch \& Lomb, FreshLook, CooperVision, Ciba Vision, United Contact Lens, Johnson \& Johnson, and ClearLab. The second model synthesizes images mimicking live irises without any textured contact lenses. 

Next, we use Neurotechnology's VeriEye SDK~\cite{neurotechnology_verieye} to identify any matches between the generated irises without TCLs and irises used to train the StyleGAN2-ADA model, and remove all synthetic samples matching live irises ({\bf Step 2} in Fig. \ref{fig:pipeline}). The proposed framework is independent of the iris matcher. We decided to use VeriEye in this work due to its availability and relatively high position in the NIST IREX program's leaderboard \cite{IREX_X_URL}.

This curated, solely-synthetic data is then used to train regular iris PAD methods ({\bf Step 3} in Fig. \ref{fig:pipeline}), which in this work are based on three popular architectures: DenseNet \cite{huang2017densely}, ResNet \cite{he2015deep}, and Vision Transformer \cite{dosovitskiy2020image}. 

Finally, the trained PAD models are tested with several state-of-the-art benchmarks composed of non-synthetic images of irises with and without TCLs ({\bf Step 4} in Fig. \ref{fig:pipeline}). Sections \ref{sec:step1} through \ref{sec:step4} provide details related to all elements of the proposed framework. Since each step accomplishes an independent task, optimizing all tasks jointly is unnecessary and was not performed.

\subsection{Training Generative Models on Authentic Data (Step 1 in Fig. \ref{fig:pipeline})}
\label{sec:step1}

\subsubsection{Datasets Used}

% UND TCL Datasets
\paragraph{Authentic TCL:} ND3D ~\cite{9164898} and ``ND Cosmetic Contact Lenses 2013'' (further: ND-CCL)~\cite{doyle2014notre} datasets were used to train the StyleGAN2-ADA model to synthesize iris images with TCLs. ND3D ~\cite{9164898} consists of 4,328 iris images captured with AD100 and LG4000 sensors, and represent three different contact lens brands: Bausch \& Lomb, FreshLook, and Johnson \& Johnson. ND-CCL 
consists of 16,925 iris images captured using the same sensors (AD100 and LG4000) and representing five different contact lens brands: CooperVision, Ciba Vision, United Contact Lens, Johnson \& Johnson, and ClearLab. In total, ND3D and ND-CCL combined provides 21,253 iris images with textured contact lenses. 

\begin{figure*}[!htbp]
\centering
  \includegraphics[width=\linewidth]{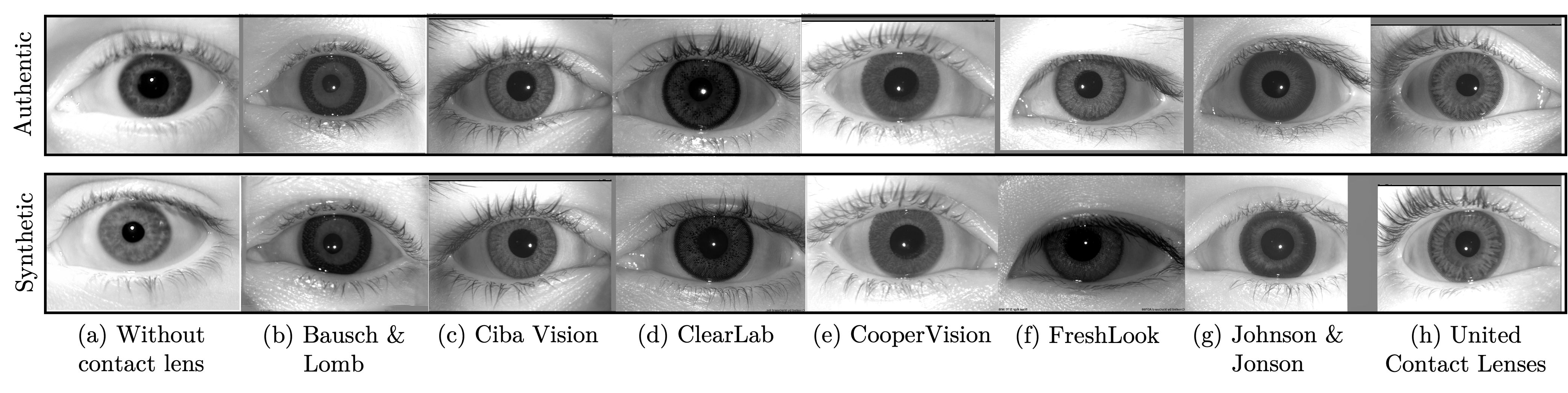}
  \caption{Examples of authentic (upper row) and synthetically-generated by a conditional StyleGAN2-ADA (bottom row) samples without textured contact lens (noTCL) and with textured contact lens (TCL) of a given brand.}
  \label{brands}
\end{figure*}  

\paragraph{Authentic noTCL:} A subset of images presenting irises without textured contact lenses, selected from all datasets published by the University of Notre Dame \cite{CVRL_DATA}, was used to train the generative model synthesizing iris images without contact lenses (noTCL). This subset contains 161,549 iris samples of an ISO-compliant resolution (640 $\times$ 480 pixels). All images were captured with the LG2200 sensor, representing 2,237 irises from 1,120 subjects.

Detailed information about the distribution of samples is presented in Table \ref{tab:dataset-refinement}. Example images from ND3D and ND-CCL are shown in the upper row in Fig. \ref{brands}. The bottom row in Fig. \ref{brands} presents example images synthesized by the models trained in Step 1, mimicking both images of irises with and without texture contact lenses.

\subsubsection{Training}

``Authentic TCL'' and ``Authentic noTCL'' collections were used to train generative models synthesizing {\bf Synthetic TCL} and {\bf Synthetic noTCL} samples (used later in Step 3), respectively. We employed a class-conditional StyleGAN2 model to generate {\bf Synthetic TCL} iris samples. For each textured contact lens brand, we defined a separate class. {\bf Synthetic noTCL} iris images were generated using an unconditional StyleGAN2 model. The training code was adopted from the NVIDIA repository\footnote{\url{https://github.com/NVlabs/stylegan3/}}, specifically using the StyleGAN2 and StyleGAN2-ADA configurations. To ensure consistent input across models, all authentic noTCL and TCL iris images were re-scaled to a standard resolution of $512\times512$ pixels (this allows the GAN model to focus on learning features independent of the original image sizes). Synthesized images were then re-scaled back to ISO-compliant resolution of $640\times480$ pixels. StyleGAN-specific training parameters\footnote{for further references, see \url{https://github.com/NVlabs/stylegan3/blob/main/docs/configs.md}} were the following: batch size = 32, gamma = 1.6384, mapping depth = 2, generator learning rate = 0.0025, and discriminator learning rate = 0.0025. Both StyleGAN-ADA models were trained from scratch using the selected iris image datasets.

\subsection{Image Synthesis and Identity Leakage Mitigation (Step 2 in Fig. \ref{fig:pipeline})}
\label{sec:step2}

\subsubsection{Synthesis} An identified issue with generative models is the potential for ``identity leakage,'' where synthetic images tend to match authentic images used in the training set. To address this issue and make our solution privacy-safe, we first generate $N>>K$ iris images, where $K$ is the number of images we eventually use in iris PAD training (Step 3). Any image that matches a sample used in StyleGAN2-ADA training is excluded. 

\subsubsection{Identity Leakage Mitigation}
Neurotechnology's VeriEye SDK~\cite{neurotechnology_verieye} was used for the leakage-related comparison experiments. This approach, however, is not matcher-specific. In our experiments $N=10,000$.

Additionally, any synthetic image that failed to enroll properly with the VeriEye was discarded. Given these exclusion rules, the remaining $K=4,167$ synthetic samples, labeled as {\bf Synthetic noTCL} in this paper, were used in Step 3. 

\subsubsection{Quality Evaluation} 
\begin{figure}[htbp]
\centering
  \includegraphics[width=1.0\columnwidth]{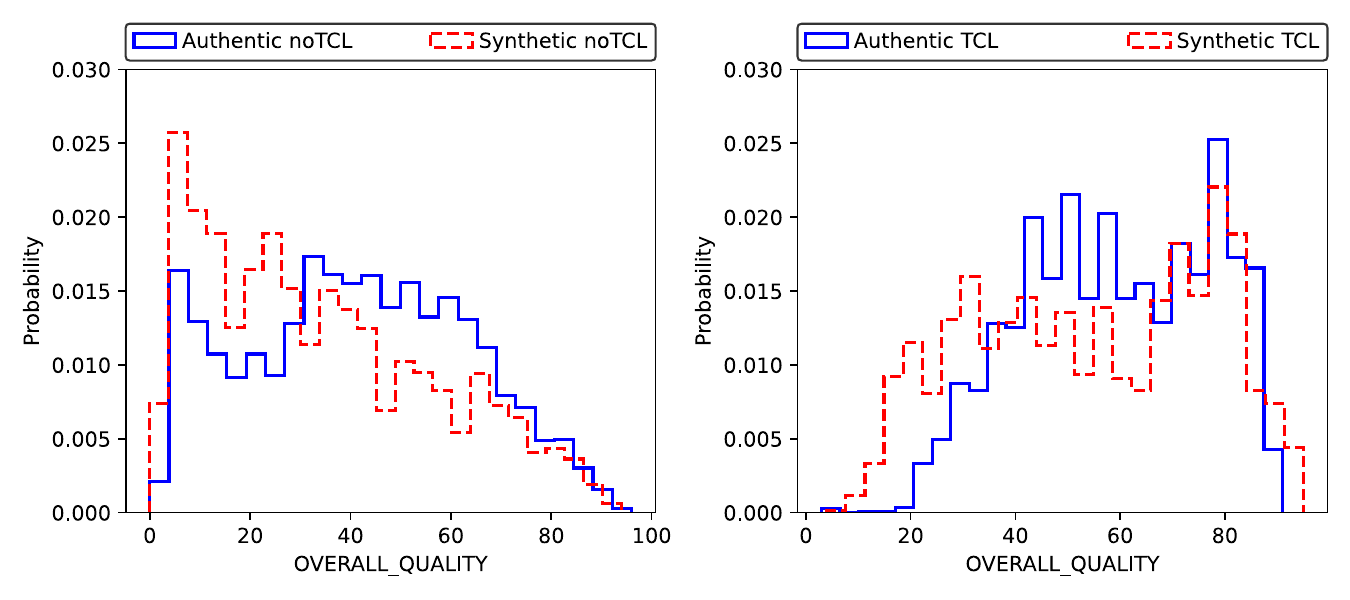}
  \caption{Distributions of the ISO/IEC 29794-6 overall quality score for iris images without contact lenses (noTCL) shown on the left plot, and for iris images with contact lenses (TCL) shown on the right plot. The quality score ranges from 0 to 100, with higher scores indicating better image quality.}
  \label{iso}
\end{figure}

We evaluate the quality of synthesized iris images using the ISO/IEC 29794-6 quality metrics \cite{iso2015iso} to ensure that synthetic TCL and noTCL images present comparable quality as authentic TCL and noTCL irises. Such agreement in quality metrics is achieved, as shown in Fig. \ref{iso}. 

%%%%%%%%%%%%%%%%%%%%%%%%%%
\begin{table*}[!htbp]

\caption{Sources of data used in training and testing of the proposed framework.}
\label{tab:dataset-refinement}
\centering
\scriptsize
\begin{tabular}{lllc} 
\toprule
\textbf{Step as in Fig. \ref{fig:pipeline}} & \textbf{Image Type} & \textbf{Contributing Datasets} & \textbf{\# of Samples}\\
\midrule

%% Step 1
Step 1 (training generative models) & Authentic TCL & Sourced from ND3D~\cite{9164898} and {ND-CCL}~\cite{doyle2014notre}: & \\ &  & ~~~~Bausch \& Lomb & 488\\ 
&  & ~~~~FreshLook & 1,901\\ 
&  & ~~~~CooperVision & 2,095\\ 
&  & ~~~~Ciba Vision & 2,716\\ 
&  & ~~~~United Contact Lens & 3,156\\ 
&  & ~~~~Johnson \& Johnson & 5,015\\ 
&  & ~~~~ClearLab & 5,863\\ 
%------------------------------------
\cline{2-4}
& Authentic noTCL & Sampled from publicly-available datasets~\cite{CVRL_DATA} & 161,549\\ 
\midrule

%% Step 2,3 (Exp 1)
Step 2 (privacy-safe synthesis), and & Synthetic TCL & Generated and offered with this paper & 4,167\\ 
%---------------------------------------
\cline{2-4}
Step 3 (PAD training with synthetic data)  & Synthetic noTCL & Generated and offered with this paper & 4,167\\ 
\midrule

%% Step 3 (Exp 2)
Step 3 (PAD training with authentic data) & Authentic TCL & Sampled from publicly-available datasets~\cite{CVRL_DATA} & 4,167\\
%---------------------------------------
\cline{2-4}
 & Authentic noTCL & Sampled from publicly-available datasets~\cite{CVRL_DATA} & 4,167\\ 
\midrule

%% Step 4 (Test Set)
Step 4 (testing with iris PAD benchmarks) & Authentic TCL & BERC\_IRIS\_FAKE \cite{Sung_OE_2007} & 140\\ 
&  & IIITD Contact Lens Iris \cite{kohli2013revisiting} & 2,256\\ 
&  & LivDet-Iris Clarkson 2015 \cite{Yambay_ISBA_2017} & 2,533\\ 
&  & LivDet-Iris Clarkson 2017 \cite{yambay2017livdet} & 1,886\\ 
&  & LivDet-Iris IIITD-WVU 2017 \cite{yambay2017livdet} & 957\\ 
%---------------------------------------------
\cline{2-4}
& Authentic noTCL & \multirow{1}{*}{BERC\_IRIS\_FAKE \cite{Sung_OE_2007}} &\multirow{1}{*}{2,733}\\ 
&  & IIITD Contact Lens Iris \cite{kohli2013revisiting} & 2,140\\ 
&  & LivDet-Iris Clarkson 2015 \cite{Yambay_ISBA_2017} & 1,884\\ 
&  & LivDet-Iris Clarkson 2017 \cite{yambay2017livdet} & 3,868\\ 
&  & LivDet-Iris IIITD-WVU 2017 \cite{yambay2017livdet} & 2,216\\ 
\bottomrule

\end{tabular}
\end{table*}

\subsection{Training of Iris PAD Algorithms with Synthetic Data Only (Step 3 in Fig. \ref{fig:pipeline})}
\label{sec:step3}

\subsubsection{Presentation Attack Detection Models} 

To investigate the performance of PAD algorithms, we adopted the DenseNet \cite{huang2017densely}, ResNet \cite{he2015deep}, and Vision Transformer (ViT) \cite{dosovitskiy2020image} architectures, which achieved a good accuracy in the most recent LivDet-Iris 2023 competition \cite{tinsley2023iris}. 

{\bf DenseNet} employs a dense connectivity pattern, where each layer receives feature maps from all preceding layers as input, and its own feature maps become inputs to all subsequent layers. This dense connectivity strengthens feature propagation, improves feature re-usability, and significantly reduces the number of parameters compared to traditional convolutional neural networks. We used a pre-trained DenseNet with 121 layers for this study.

{\bf ResNet}, instead of directly feeding information from layer to layer, introduces skip connections that bypass intermediate layers, allowing information to flow directly from earlier layer (L) to later layer (L+2). This enables the model to learn the difference between the two representations, rather than starting from scratch at each layer. This approach allows the model to build upon existing knowledge, facilitating the learning of complex and fine-grained features, and ultimately leading to improved performance in both shallow and deep networks. Both DenseNet and ResNet models tackle the problem of vanishing gradients, a fundamental challenge in training deep networks. We used a pre-trained ResNet with 101 layers for this study.

{\bf Vision Transformer (ViT)}, instead of feeding the entire image at once, it breaks the input down into smaller, fixed-size patches. These patches are then linearly embedded along with their position and passed to the transformer encoder, which analyzes the relationships between individual patches and the entire image, and classifies the image based on these learned representations. We used a pre-trained ViT-base model with a patch size of 16 for this study.

\subsubsection{Training}

In this step, the PAD models are trained solely on synthetic data. Although the ISO quality metrics are similar to those obtained for authentic iris images (cf. \ref{iso}), to improve the model's robustness to image quality variations and enhance the diversity of the iris data during training, we employed various image augmentation techniques inspired by the \verb+imgaug+ library \cite{imgaug}. These techniques included basic transformations (\eg, flipping, rotation), additive noise, and more advanced distortions like sharpening, blurring, and brightening. To enhance the similarity between synthetic and authentic irises, we augmented the synthetic irises used for training process of experiment 1 with a more rigorous brightness and contrast adjustment strategy and a broader range of blurring techniques.

Since there are 4,167 synthetic noTCL privacy-safe samples (curated in Step 2) available in this step, we included an equal number of synthetic TCL samples (4,167) representing seven contact lens types, with 595 or 596 samples per lens brand to achieve well-balanced set across all brands. 80\% of the data was used for training, with the remaining 20\% reserved for validation. Each model was trained for a maximum of 50 epochs, and the weights corresponding to the best accuracy on the validation set were picked for the final model. A batch size of 32 was used with stochastic gradient descent (SGD) as the optimizer and cross-entropy as the loss function. The learning rate was 0.005, the weight decay was 1e-6, and the momentum was 0.9.

\subsection{Testing of Iris PAD Algorithms with Unseen Authentic Data (Step 4 in Fig. \ref{fig:pipeline})}
\label{sec:step4}

To evaluate the effectiveness of the PAD models trained on synthetic data and compare them to PAD models trained traditionally (with authentic data), %in distinguishing noTCL and TCL irises, 
we utilized several standard iris PAD benchmarks containing both noTCL and TCL authentic samples: BERC\_IRIS\_FAKE \cite{Sung_OE_2007}, IIITD Contact Lens Iris \cite{kohli2013revisiting}, LivDet-Iris Clarkson 2015 \cite{Yambay_ISBA_2017}, LivDet-Iris Clarkson 2017 \cite{yambay2017livdet}, and LivDet-Iris IIITD-WVU 2017 \cite{yambay2017livdet}. To our knowledge, this composition of benchmarks is the largest one can presently collect from publicly-available research datasets containing images with textured contact lenses. In the interest of fair evaluation, we have excluded textured contact lens datasets acquired by the University of Notre Dame, since data from that group was used at various stages of the method's design. For consistency, all images were center-cropped (using circular iris boundary approximations obtained by software offered with \cite{trokielewicz2020post}) and resized to a uniform resolution of $256\times256$ pixels, which allows the models to be more focused on the iris texture. 
Details of the datasets used in various steps when designing the proposed framework are provided in Table \ref{tab:dataset-refinement}.

\section{Experiments and Results}

\subsection{Cross-validation and Metrics}

Following ISO/IEC 30107-3:2017 \cite{ISO_IEC_301073:2017}, we evaluate PAD models by reporting Attack Presentation Classification Error Rate (APCER: the proportion of ``attack'' samples called ``bonafide'') and Bonafide Presentation Classification Error Rate (BPCER: the proportion of ``bonafide'' samples called ``attack''). APCER and BPCER are used to (a) plot Detection Error Tradeoff (DET) curve, (b) calculate Area Under the Receiver Operating Characteristic curve (AUROC), and (c) calculate the decidability score: 

$$
d' = \frac{| \mu_{\textrm{BF}} - \mu_{\textrm{PA}} |}{\sqrt{0.5\big(\sigma^2_{\textrm{BF}} - \sigma^2_{\textrm{PA}}\big)}}
$$
\noindent
where $\mu$ and $\sigma$ are means and standard deviations, respectively, of the PAD scores obtained for bona fide (BF) and attack (PA) samples.

For estimating the uncertainty of the results associated with stochastic optimization during neural network models training, we independently trained each model five times with the same datasets, but with five different training seeds. We thus further report the average performance metrics along with their standard deviations.

\subsection{Experiments}

To evaluate a potential gap between the performance of the iris PAD methods trained traditionally (with authentic samples) and trained solely with synthetic samples, we designed two experiments:

\begin{enumerate}
    \item[{\bf E1:}] the iris PAD models were {\bf trained exclusively on synthetic data}, as described in Sec. \ref{sec:step3}, and {\bf tested on iris PAD benchmarks},
    \item[{\bf E2:}] the iris PAD models were {\bf trained solely on authentic data} (including images of irises with and without textured contact lenses), and {\bf tested on the iris PAD benchmarks}; to ensure fair comparisons, we sampled 4,167 authentic iris images without textured contact lenses (noTCL), and 4,167 authentic iris images with textured contact (TCL) for training, representing all identities whose data was used also for training generative models in Step 1. 
\end{enumerate}

\subsection{Results}
\label{sec:4.3}

\begin{figure*}[!ht]
\centering
\begin{subfigure}{0.32\linewidth}
    \includegraphics[width=\textwidth]{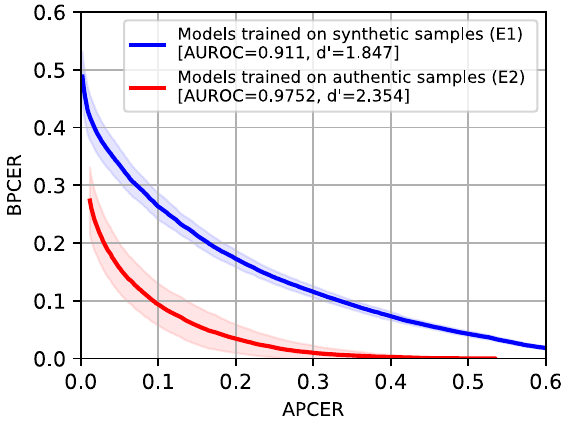}
    \caption{DenseNet}
\end{subfigure}
\hfill
\begin{subfigure}{0.32\linewidth}
    \includegraphics[width=\textwidth]{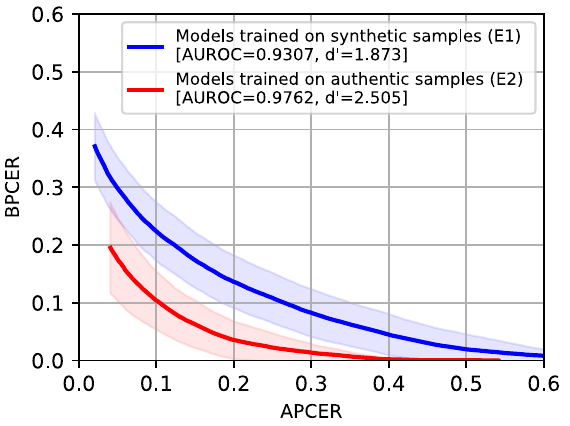}
    \caption{ResNet}
\end{subfigure}
\hfill
\begin{subfigure}{0.32\linewidth}
    \includegraphics[width=\textwidth]{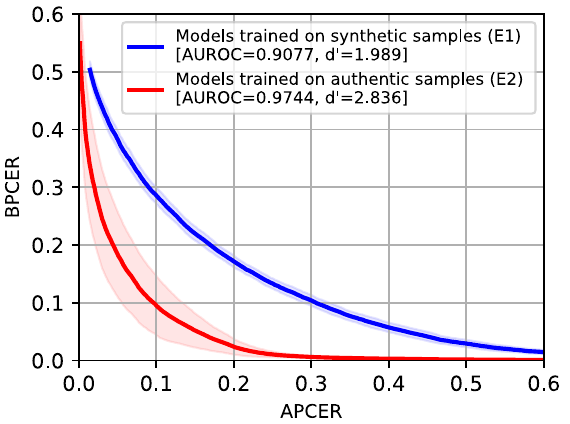}
    \caption{ViT}
\end{subfigure}

\caption{Average DET curves (thick lines), with shaded areas representing one standard deviation from five train-test runs, obtained for both experiments (E1 in blue and E2 in red) and for all three model backbones.}
\label{det_curve}
\end{figure*}

\begin{table}[!b]
\centering
%\scriptsize
\caption{BPCER (in \% and averaged across five train-test runs) at different APCER levels obtained for iris PAD methods utilizing three architectures. \textbf{E1:} models trained solely with synthetic samples. \textbf{E2:} models trained solely with authentic samples.}
\label{tab:bpcer}
\scriptsize
\tabcolsep=0.11cm
\begin{tabular}{lcccccc}
\toprule
\multirow{2}{*}{} & \multicolumn{2}{c}{\textbf{DenseNet}} & \multicolumn{2}{c}{\textbf{ResNet}} & \multicolumn{2}{c}{\textbf{ViT}} \\ \cline{2-7}
APCER& \multicolumn{1}{c}{E1} & \multicolumn{1}{c}{E2} & 
\multicolumn{1}{c}{E1} & \multicolumn{1}{c}{E2} &
\multicolumn{1}{c}{E1} & \multicolumn{1}{c}{E2} \\ \midrule
0.1\% & $53.17$ & $33.08$ & $44.85$ & $34.79$ & $53.93$ & $56.86$ \\ 
1.0\%   & $42.72$ & $25.03$ & $39.34$ & $24.66$ & $47.30$ & $35.57$ \\ 
5.0\%   & $33.60$ & $14.20$ & $28.68$ & $14.04$ & $35.50$ & $16.56$ \\ 
10.0\%  & $26.41$ & $8.52$  & $21.6$  & $8.41$  & $27.27$ & $8.52$ \\ 
\bottomrule

\end{tabular}
\end{table}

\begin{figure*}[!htbp]
\centering
\includegraphics[width=1.0\linewidth]{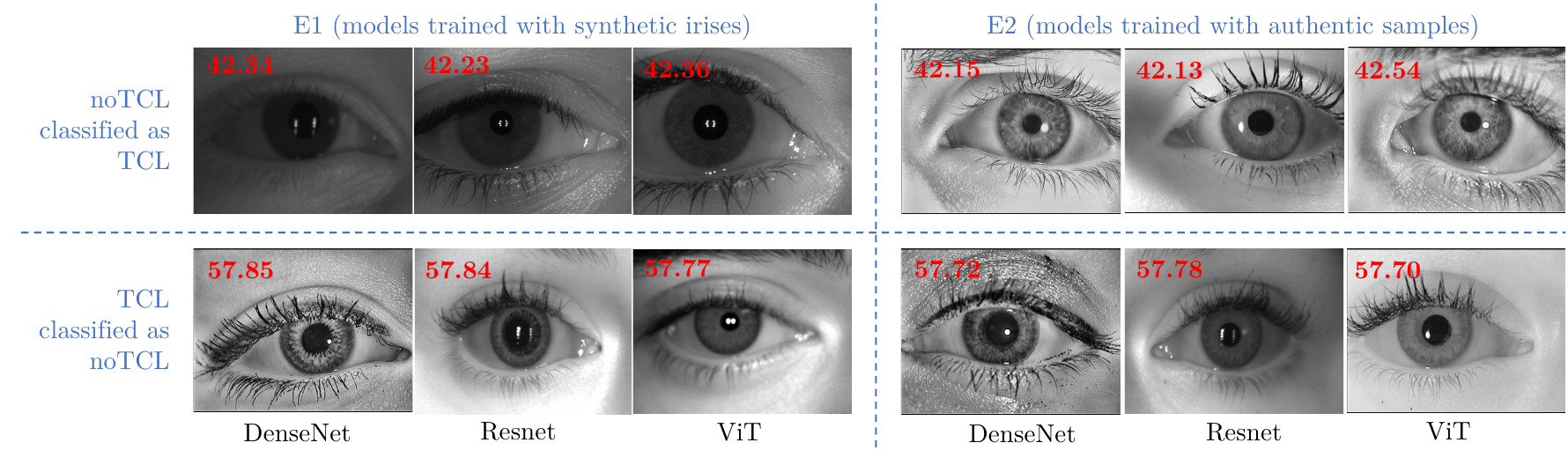}
  \caption{Examples of noTCL samples classified as TCL, and TCL samples classified as noTCL for each model. Numbers over each image is the liveness detection score. The score closer to 100.0 indicates clean iris (noTCL sample), and the score closer to 0.0 indicates an iris with textured contact lens (TCL). A threshold on 50.0 was used in classification into TCL and noTCL.}
  \label{misclassified_samples}
\end{figure*}  

Figure~\ref{det_curve} shows detection error tradeoff (DET) curve averaged across five train-test runs for all model architectures. These results are obtained on the iris PAD benchmarks (as described in Table~\ref{tab:dataset-refinement}) in both experiments (E1 and E2). Additionally, Table ~\ref{tab:bpcer} shows a few selected operational points on the DET curves, specifically the average BPCER at various APCER values.
It can be clearly seen that models trained exclusively with authentic samples (E1) outperformed models trained traditionally with synthetic iris data (E2), although the gap between two is not significant. For instance, the average AUROC for models trained with authentic data reached 97\%, while the average AUROC values for models trained with synthetic data ranged from 90\% (lowest) to 93\% (highest), with the lowest and highest values obtained by the DenseNet and ResNet-based models, respectively. This performance gap, however, is not significant, and the entire experiment brings an interesting observation that iris PAD methods can be trained solely on relatively small, synthetically-generated datasets and still achieve satisfactory performance. Note that experiment E1 assumed using only privacy-safe synthetic data. The number of training samples (two sets of only 4,167 images each) was selected intentionally to run comparisons with models trained on authentic data, which were also limited to two sets of 4,167 images. 

One potential reason for the performance gap is that synthetic samples may lack the diversity in term of brightness, contrast, blurriness, and other factors related to image quality, which are present in authentic samples. Additionally, the use of data from a single sensor in training the generative model in Step 1 might have further limited the diversity of synthetic images. 
Figure~\ref{misclassified_samples} illustrates examples of wrong classifications, likely caused by differences in image quality factors between the train and test sets. For instance, models trained with synthetic samples (on which we are focusing in this paper) failed to classify samples that are too dark (upper left part of Fig.~\ref{misclassified_samples}) or those that were either too blurry or too sharp (lower left part of Fig.~\ref{misclassified_samples}). This may suggest what additional augmentations could be made to increase the diversity of synthetic training samples. Since the tests presented in Step 4 are made on a sequestered set of benchmarks, we certainly did not make any adjustments in our augmentation strategies after seeing these mis-classifications.

To assess whether there are performance differences between model architectures (DenseNet, ResNet and ViT) we conducted paired samples t-test, comparing mean AUROC values separately for (a) models trained with authentic (red curves in Fig. \ref{det_curve}) and (b) synthetic (blue curves in Fig. \ref{det_curve}) iris images. In each (a) and (b) scenario, we conducted three tests comparing AUROC values of two models, namely: DenseNet vs ResNet, DenseNet vs ViT, and ResNet vs ViT. The null hypothesis was that there are no differences between mean values. The $p$-values in scenario (a) ranged from 0.75 to 0.82, while those in scenario (b) ranged from 0.17 to 0.61. Thus, there are no reasons, at the assumed significance level $\alpha=0.05$, to reject the null hypothesis amd hence we conclude that there are no statistically significant differences in the performance across model architectures.

\section{Conclusions}
This paper proposed the framework in which exclusively synthetically-generated iris images were used to build the entire iris PAD method detecting textured contact lenses. This study demonstrates that it is possible to train effective iris PAD models without using any authentic data, collected from human subjects. To achieve this goal we trained unconditional generative models synthesizing iris images without contact lenses, and conditional generative models synthesizing images of irises wearing contact lenses offered by seven different manufacturers. By applying an ``identity leakage'' mitigation mechanism in the pipeline, the proposed framework offers an advantage of reducing privacy concerns associated with using iris data from authentic subjects. As a result, we obtained privacy-safe iris PAD methods that perform comparably well when tested on all the existing benchmarks offering iris images with and without textured contact lenses (benchmarks used in models training were excluded from testing to avoid bias).

\vskip2mm\noindent
{\bf Possible extensions of this work:} One obvious extension of this work is to keep generating synthetic data and see when the performance gap is filled out (if at all). The second extension of this work is to mix the existing authentic datasets used to train iris PAD to date with synthetically-generated samples (both mimicking irises not covered by textured contact lenses, and images of irises wearing contact lenses).

\vskip2mm\noindent
{\bf Acknowledgments:} This material is based upon work partially supported by the National Science Foundation under Grant No. 2237880. Any opinions, findings, and conclusions or recommendations expressed in this material are those of the authors and do not necessarily reflect the views of the National Science Foundation.

%%%%%%%%%%%%%%%%%%%%%%%%%%%%%%%%%%%%%%%%%%%%%%%%%%%
{\small
\bibliographystyle{ieee}
\bibliography{egbib}
}
\end{document}